\documentclass[letterpaper, conference]{cssconf}
\IEEEoverridecommandlockouts   
\usepackage{hyperref}       
\usepackage{url}            
\usepackage{booktabs}       
\usepackage{amsfonts}       
\usepackage{nicefrac}       
\usepackage{microtype}      
\usepackage{xcolor}         
\usepackage{graphicx}
\usepackage{amsmath}
\usepackage{subcaption}

\title{On the Design of Safe Continual RL Methods for Control of Nonlinear Systems}

\author{Austin Coursey, Marcos Quinones-Grueiro, and Gautam Biswas
\thanks{*This material is based upon work supported by the National Science Foundation Graduate Research Fellowship Program under Grant No. 2444112. Any opinions, findings, and conclusions or recommendations expressed in this material are those of the author(s) and do not necessarily reflect the views of the National Science Foundation.}
\thanks{Institute for Software Integrated Systems,
        Vanderbilt University, Nashville, TN, USA
        {\tt\small austin.c.coursey@vanderbilt.edu}}%
}

\begin{document}
\maketitle
\thispagestyle{empty}
\pagestyle{empty}

\begin{abstract}

Reinforcement learning (RL) algorithms have been successfully applied to control tasks associated with unmanned aerial vehicles and robotics. In recent years, safe RL has been proposed to allow the safe execution of RL algorithms in industrial and mission-critical systems that operate in closed loops. However, if the system operating conditions change, such as when an unknown fault occurs in the system, typical safe RL algorithms are unable to adapt while retaining past knowledge. Continual reinforcement learning algorithms have been proposed to address this issue. However, the impact of continual adaptation on the system's safety is an understudied problem. In this paper, we study the intersection of safe and continual RL. First, we empirically demonstrate that a popular continual RL algorithm, online elastic weight consolidation, is unable to satisfy safety constraints in non-linear systems subject to varying operating conditions. Specifically, we study the MuJoCo HalfCheetah and Ant environments with velocity constraints and sudden joint loss non-stationarity. Then, we show that an agent trained using constrained policy optimization, a safe RL algorithm, experiences catastrophic forgetting in continual learning settings. With this in mind, we explore a simple reward-shaping method to ensure that elastic weight consolidation prioritizes remembering both safety and task performance for safety-constrained, non-linear, and non-stationary dynamical systems. 

\end{abstract}

\maketitle

\section{Introduction} \label{sec:intro}

Deep reinforcement learning (RL) algorithms have shown recent success in a variety of control applications. These include unmanned aerial vehicle attitude control \cite{koch2019reinforcement}, simulated racecar driving \cite{wurman2022outracing}, and robotics tasks \cite{apolinarska2021robotic}. When controlling real systems, we must satisfy safety constraints, especially in safety-critical applications. Despite their success, a notable challenge of RL algorithms is maintaining safety, limiting their real-world use \cite{dulac2021challenges}.  

The field of safe reinforcement learning has emerged to address this challenge. Safe RL approaches can be broadly divided into model-based and model-free approaches \cite{gu2022review}. Model-based approaches include model-predictive control \cite{zanon2020safe}, methods that use Lyapunov functions \cite{berkenkamp2017safe} or control barrier functions \cite{marvi2021cbf} to guarantee safety to some probability \cite{gu2022review} and methods that use formal verification \cite{alshiekh2018sheild} to never violate safety. The model used in these methods may be a known physical model \cite{zanon2020safe} or a derived data-driven model (e.g., in \cite{lutjens2019safe} they learn collision probabilities). The effectiveness of model-based safe RL algorithms depends on the model's accuracy. When a system lacks an available accurate dynamics model, model-free safe RL approaches may be required. Many popular model-free safe RL approaches perform constrained optimization. Examples of these include constrained policy optimization \cite{achiam2017constrained} which has guarantees on near-constraint satisfaction and Lagrangian proximal policy optimization (PPO-Lag) \cite{ray2019benchmarking}. While PPO-Lag and CPO show comparable performance, CPO has been shown to lead to more stable safety satisfaction \cite{ji2023safety}. Despite the rapid growth of safe RL, the performance of safe RL algorithms in non-stationary environments is understudied. As we control a system over its lifetime, the environment will change. These changes could be caused by component degradations leading to sudden faults, requirement shifts, or unknown environmental encounters. An early work \cite{ammar2015safe} developed a policy gradient method for safe lifelong RL, ignoring the challenge of catastrophic forgetting. Some recent works study safety in non-stationary meta-learning RL environments \cite{chen2021context}, but there is still a large gap in the study of safety in online, lifelong non-stationarity. 

The fields of continual reinforcement learning and lifelong reinforcement learning aim to adapt to task and environment changes over the lifetime of a system. The key measures of success for a continual RL algorithm are the ability to avoid catastrophic forgetting and the forward and backward transfer across environment changes \cite{khetarpal2022towards}. Continual RL approaches are largely regularization-based, use experience replays or knowledge bases, or perform network expansion. Regularization-based approaches, such as elastic weight consolidation \cite{kirkpatrick2017overcoming}, add a penalty to the reward function that encourages the network to remember how to operate in previously seen conditions.  Replay-based methods, such as CLEAR \cite{rolnick2019experience}, encourage long-term memory in the replay buffer used in off-policy RL methods. Knowledge base methods, such as \cite{zhan2017scalable}, take a similar approach to ensure the knowledge base contains relevant information from all conditions. Expansion-based approaches, such as \cite{zhang2023dynamics}, expand a part of the network each time a new scenario is encountered. This expansion may be on the network level \cite{kessler2022same} or through mixture models \cite{xu2020task}. In that way, the parts of the network that were optimal on previously seen scenarios are never overwritten, but this introduces scalability concerns. None of these approaches account for the safety of the system. The intersection of safe and continual RL is an understudied and open problem that is highly important as we work toward more real-world applications of reinforcement learning.

With this in mind, in this paper, we empirically demonstrate the need for safe continual RL algorithms. We focus on problems with sudden, dramatic changes in the system while learning like when a fault occurs or equipment breaks. These are simulated by removing joints in the HalfCheetah and Ant MuJoCo environments. In these, we want the agent to continually learn to improve reward while satisfying a maximum velocity safety constraint. When the joint breaks off, the agent should not forget how to control the nominal robotic system, as the system will be repaired. We demonstrate that agents trained using constrained policy optimization (CPO) \cite{achiam2017constrained}, a safe RL algorithm, maintain minor cost violations across the lifetime of these agents but catastrophically forgets prior performance. We show that adding elastic weight consolidation (EWC) \cite{kirkpatrick2017overcoming} to the learning process of an agent trained with proximal policy optimization (PPO) (such as in \cite{nath2023sharing}), a continual RL algorithm (PPO+EWC), has less catastrophic forgetting than CPO but heavily violates safety. We show that a simple reward-shaping approach that penalizes PPO+EWC for safety violations, which we call \textbf{Safe EWC}, exhibits less catastrophic forgetting than CPO while maintaining comparable safety constraint satisfaction.

In this work, we make the following contributions.

\begin{enumerate}
    \item We empirically demonstrate the need for safe, continual reinforcement learning algorithms. We show that agents trained with CPO, a safe RL algorithm, experience higher catastrophic forgetting and less backward transfer than those trained with PPO+EWC, a continual RL algorithm. At the same time, agents trained with PPO+EWC ignore safety to maximize reward. 
    \item We demonstrate how a simple reward shaping modification to PPO+EWC can improve safety in the HalfCheetah and Ant MuJoCo environments that are constrained by a maximum velocity. At the same time, this modified algorithm, which we call \textbf{Safe EWC}, still reduces catastrophic forgetting and improves backward transfer.
\end{enumerate}

Code for this paper is available at \url{https://github.com/MACS-Research-Lab/safe-continual}.
\section{Problem Definitions and Assumptions} \label{sec:background}

Reinforcement learning aims to find a solution to a \textbf{Markov Decision Process (MDP)}. An MDP is a tuple consisting of a state space $\mathcal{S}$, an action space $\mathcal{A}$, a transition function $T: \mathcal{S} \times \mathcal{A} \to \mathcal{S}$, a reward function $r: \mathcal{S} \times \mathcal{A} \to \mathbb{R}$, and a discount factor $\gamma \in [0, 1)$. The MDP $=(\mathcal{S}, \mathcal{A}, T, r, \gamma)$. The goal of RL is to find a policy $\pi: \mathcal{S} \to \mathcal{A}$ that maximizes the expected discounted reward.

\begin{equation}
    G_t = \sum_{k=0}^\infty \gamma^k r(s_{t+k}, a_{t+k})
\end{equation}

Therefore, the objective is to find an optimal policy $\pi^*$, one that satisfies the objective below.

\begin{equation} \label{eq:safe}
    \pi^* = \arg\max _\pi \mathbb{E}[G_t | \pi]
\end{equation}

To introduce the notion of safety, we can naturally extend the MDP to a \textbf{Constrained Markov Decision Process (CMDP)} \cite{altman1999constrained}. In a CMDP, we also introduce cost functions $\mathcal{C}: \mathcal{S} \times \mathcal{A} \to \mathbb{R}^m$ that are each constrained by a maximum cost $d=(d_1, d_2, \dots, d_m)$. The tuple is then $=(\mathcal{S}, \mathcal{A}, T, \mathcal{C}, r, \gamma, d)$. These are typically used to constrain the objective from Equation \ref{eq:safe} as follows.

\begin{equation} \label{eq:cmdp-primal}
    \text{s.t. } \mathbb{E}[\sum_{k=0}^\infty \gamma^k \mathcal{C}_i(s_{t+k}, a_{t+k})] \le d_i \text{ for } i=1,2,\dots, m
\end{equation} 

In this formulation, we constrain the expectation of sum of the discounted cost to be below some maximum value. In scenarios with hard safety constraints, safety should never be violated. In this paper, we focus on soft safety constraints where we wish to minimize the number of safety violations, but they may occur.

A continual reinforcement learning problem can be characterized by a \textbf{Non-stationary Markov Decision Process (NSMDP)}. An NSMDP is a set $\{\mathcal{S}, \mathcal{T}, \mathcal{A}, T(s' | s, a, t), (r_t)_{t\in\mathcal{T}}, \gamma\}$, where the transition function and reward are now dependent on a set of decision epochs (time), $\mathcal{T}=\{1, 2, \dots,\infty\}$. In other words, the environment can change over time. An ideal solution to this NSMDP should avoid \textbf{catastrophic forgetting}. That is, if the environment returns to a previously seen environment, the agent should remember its policy. Additional properties such as exhibiting positive forward or backward transfer are also desirable \cite{khetarpal2022towards}. \textbf{Forward transfer} measures how well learning on one set of environmental conditions improves learning on another. \textbf{Backward transfer} measures how well learning on new environmental conditions improves learning on old environmental conditions.

In this paper, we focus on sudden, drastic changes in the environment. We refer to each instance of one of these environments as a \textbf{task}. For example, controlling an octocopter drone is one task. If a motor fault occurs, controlling an octocopter with 7 functioning motors is another task. For simplicity, we assume we know when a task change occurs. We make this assumption to assess the performance of the ``continual'' aspect of the continual RL algorithm and not the accuracy of a separate task detection block. In practice, fault detection and isolation algorithms could be used for task detection in the scenarios studied in this paper.

With these definitions established, we can define a \textbf{safe continual learning problem} as one that aims to solve a Non-stationary Constrained Markov Decision Process (NSC-MDP). NSC-MDP $= \{\mathcal{S}, \mathcal{T}, \mathcal{A}, \mathcal{C}, T(s' | s, a, t), (r_t)_{t\in\mathcal{T}}, \gamma\}$. The objective is then still Equation \ref{eq:safe}, except with the reward in $G$ time-dependent, constrained by Equation \ref{eq:cmdp-primal}. Notice that the cost is not a function of time. We assume that the safety constraint is fixed across tasks. However, for some systems, the definition of safety may also change as the task does. 
\section{Method} \label{sec:method}

\begin{figure*}[t]
    \centering
    \includegraphics[width=2\columnwidth]{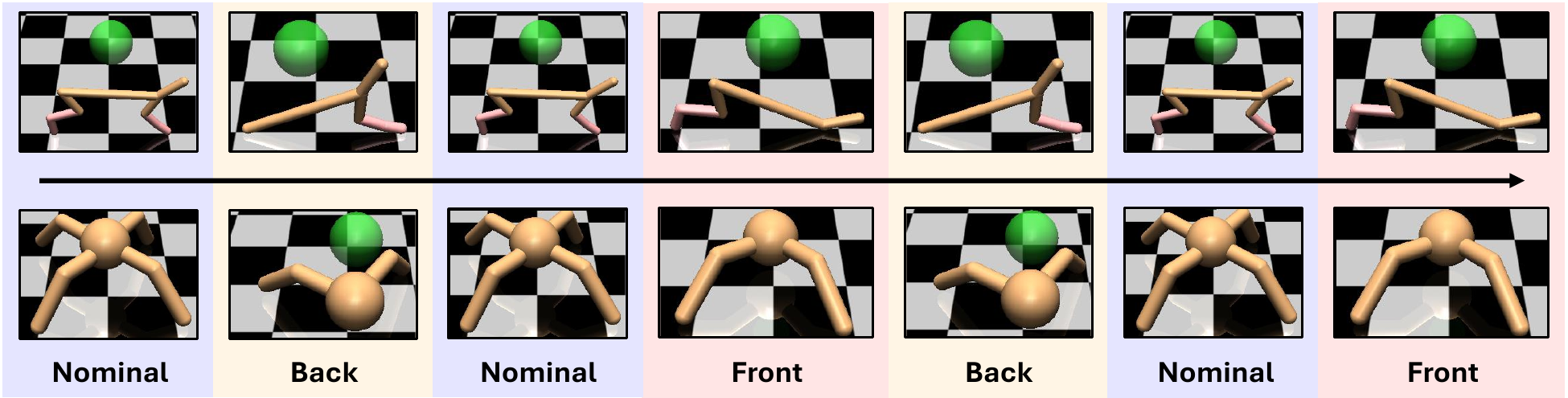}
    \caption{Task sequence for safe continual reinforcement learning. The top sequence is the MuJoCo HalfCheetah. The bottom is the Ant. Task changes occur every 1 million training timesteps and the cycle repeats. The tasks are designed to replicate a challenging and drastic change in operating mode caused by equipment being repaired or suddenly breaking due to physical damage or a fault. The objective for the environments is to travel as far as possible in a fixed amount of time while maintaining velocity constrained (visualized by the green bubble).}
    \label{fig:task_sequence}
\end{figure*}

\subsection{Control Tasks}

To perform empirical studies on safe continual RL, we need non-stationary environments with safety constraints. To create these, we modified the velocity-constrained MuJoCo benchmark included in the Safety Gymnasium \cite{ji2023safety} Python library. These include the HalfCheetah and Ant locomotion environments. In these environments, the robot needs to travel as far as possible. However, the robot is given a velocity constraint. Going above the fixed velocity threshold is dangerous, as it risks the robot's safety. Therefore, the objective and cost conflict, making this a challenging task.

For the HalfCheetah, the observation space is a 17-dimensional vector consisting of angles, velocities, and angular velocities of its body parts along with the Z position of its front tip. There are 6 rotors, one on each thigh, shin, and foot, that make up the action space. The reward for the HalfCheetah is the forward progress it makes penalized by the control costs. For the Ant, the observation space is a 105-dimensional vector consisting of the positions of the body parts, the velocities of the body parts, and the center of mass based external forces on the body parts. The action space includes the 8 torques that can be applied to the hinge joints. Its reward is the same as the cheetah, with an added reward of 1 for every timestep the ant is healthy, meaning its torso height is too high or low, and an additional penalty if the external contact forces are too high.

To make these safe RL environments non-stationary, we emulated a sudden fault or equipment damage every 1 million timesteps. We also perform maintenance between each of the 1 million timestep missions to return back to a nominal state. The sudden damage takes the form of joints being broken off. In the HalfCheetah, we remove the front or back leg. In the Ant, we remove both the front or back legs. The task sequence we designed is shown in Fig. \ref{fig:task_sequence}. Each task is revisited at least once and the nominal task is experienced most frequently. We designed this so that the optimal policy in each task would need to be different. For example, if the back leg is missing, the agent needs to crawl instead of walk. We observed that in easier scenarios, such as with parameter changes like mass or friction, the agent could learn a single policy that solves all tasks. In these cases, continual RL algorithms are not required.

\subsection{Algorithm Details}

The first goal of this paper is to determine how well agents trained with a safe RL algorithm perform in a continual RL setting. We chose constrained policy optimization (CPO) \cite{achiam2017constrained} as the representative safe RL algorithm. We selected this algorithm because it has been shown to be effective in improving safety in scenarios with soft safety constraints, and is more stable than Lagrangian proximal policy optimization \cite{ji2023safety}. CPO handles constraints directly in the policy optimization process. It performs trust-region policy updates that ensure the policy at the next step is not outside a stable region where behavior may differ dramatically. It defines a policy update for CMDPs that guarantees both cost satisfaction and increases in reward. The CPO policy update is defined as follows \cite{achiam2017constrained} where $A$ is the advantage function, $d^\pi$ is the discounted future state distribution, $J$ is the expected discounted future return, $\delta > 0$ is the step size, and $\bar{D}_{KL}$ is the KL divergence used to measure divergence between policies. 

\begin{align*}
\pi_{k+1} =&\arg \max_{\pi \in \Pi_{\theta}} \underset{s \sim d^{\pi_k}, a \sim \pi}{\mathbb{E}} \left[ A^{\pi_k}(s, a) \right] \\
\text{s.t.} \quad &J_{C_i}(\pi_k) + \frac{1}{1 - \gamma} \underset{s \sim d^{\pi_k}, a \sim \pi}{\mathbb{E}} \left[ A^{\pi_k}_{C_i}(s, a) \right] \leq d_i \quad \forall i \\
&\bar{D}_{KL}(\pi || \pi_k) \leq \delta.
\end{align*}

We used the CPO implementation from the SafePO \cite{ji2023safety} Python library since it has been validated on the MuJoCo safe velocity tasks and comes with optimized hyperparameters on these environments, saving us significant computational effort.  

Next, we want to determine how well agents trained with a continual RL algorithm maintain safety. We chose proximal policy optimization (PPO) with elastic weight consolidation (EWC) \cite{kirkpatrick2017overcoming} as the continual RL algorithm in this paper. EWC is a popular continual learning algorithm. The key idea behind EWC is to determine which weights of the neural network were most important for solving the previous task. Then, the network is penalized for drastically changing those weights. This penalty is enforced on the neural network loss, shown in the following equation \cite{kirkpatrick2017overcoming}, where $\theta$ are neural network parameters, $\theta^*_A$ are the optimal parameters for the previous task, $\mathcal{L}_B$ is the loss on the current task, $F$ is the approximated Fisher information matrix that measures how important each parameter was to the previous task, $i$ is the parameter index, and $\lambda$ is a hyperparameter that determines the tradeoff between remembering previous tasks and learning new ones.

\begin{equation}
\mathcal{L}_{\text{EWC}}(\theta) = \mathcal{L}_{\text{B}}(\theta) + \frac{\lambda}{2} \sum_{i} F_{i} \left( \theta_{i} - \theta_{A, i}^{*} \right)^2.
\end{equation}

To incorporate EWC with PPO, a powerful reinforcement learning algorithm, we apply the EWC loss to the neural network in PPO that makes actions, the actor. We calculate the approximate Fisher information matrix using the final 20 episodes of observations on a task. We save a separate Fisher information matrix for each task in the task sequence to ensure that all tasks can be remembered. Then, we calculate each task's EWC loss independently and sum them into a single EWC penalty.

Choosing a reasonable $\lambda$ is an important step to ensure EWC works properly. To determine a reasonable $\lambda$ for our experiments, we ran PPO+EWC on the HalfCheetah with the task sequence $\{\text{nominal}, \text{back}\}$. At each epoch, we evaluated the agent on both tasks. An agent with the best nominal reward at the end of this sequence remembers the best, and an agent with the best back reward at the end of this sequence learns the second task the best. We ran this experiment for a grid of $\lambda=\{0.5, 1, 5, 10, 25, 100\}$. We found that using any $\lambda > 0$ (using PPO+EWC instead of just PPO) led to less forgetting. However, increasing $\lambda$ did not necessarily lead to more remembering, likely due to the nature of the complex, multi-objective optimization. We observed that $\lambda=10$ led to the most stable nominal performance while learning on the back task and had similar learning abilities on the back task. Therefore, we chose $\lambda=10$.

To incorporate safety into a continual RL algorithm, we take a simple reward-shaping approach. We call this approach Safe EWC. We modify the original task reward to be discounted by the safety violations. Therefore, the new reward is as follows, where $\beta$ is a cost weight hyperparameter.

\begin{equation}
    r_{\text{Safe EWC}}(s, a) = r(s, a) - \beta C(s, a)
\end{equation}

By shaping the reward with the cost, we are effectively adding another penalty to the PPO+EWC loss function. This penalty encourages safe behavior. However, the performance of this may heavily depend on the value of $\beta$. An improperly chosen $\beta$ could cause the agent to ignore safety or focus too heavily on safety. Though it is worth noting we did not experience this problem. The advantage of an algorithm like CPO is it ``automatically picks penalty coefficients to attain the desired trade-off between reward and constraint cost'' \cite{achiam2017constrained}. Therefore, Safe EWC is a first step at a safe continual reinforcement learning algorithm, but more sophisticated algorithms that are less dependent on hyperparameters should be developed in future work. 

To choose the $\beta$ coefficient for our problem, we divided the maximum reward achieved by PPO+EWC with the maximum cost to make the cost as important as the reward. This gave us $\beta=5$.

\section{Results} \label{sec:results}

With the algorithms established and hyperparameters determined, we trained agents using each of the three algorithms on the task sequence shown in Fig. \ref{fig:task_sequence}. We also returned back to the nominal task at the end, allowing 8 million total training interactions in the sequence. To account for randomness in the training process, we ran each algorithm with 5 different seeds. For each seed, we parallelized the training process to train in 10 parallel threads. Each seed was also trained in parallel, meaning 50 instances of the environment were run at once. This was done on an AMD Ryzen Threadripper 3960X 24-Core CPU.

\subsection{Case Study 1: HalfCheetah}

\begin{figure}[t]
    \centering
    \begin{subfigure}[b]{\columnwidth}
        \centering
        \includegraphics[width=\columnwidth]{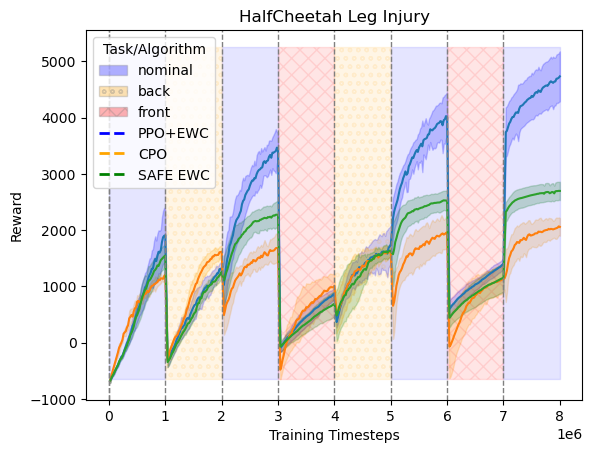} 
        \caption{Training reward curves.}
        \label{fig:cheetah_reward}
    \end{subfigure}

    \begin{subfigure}[b]{\columnwidth}
        \centering
        \includegraphics[width=\columnwidth]{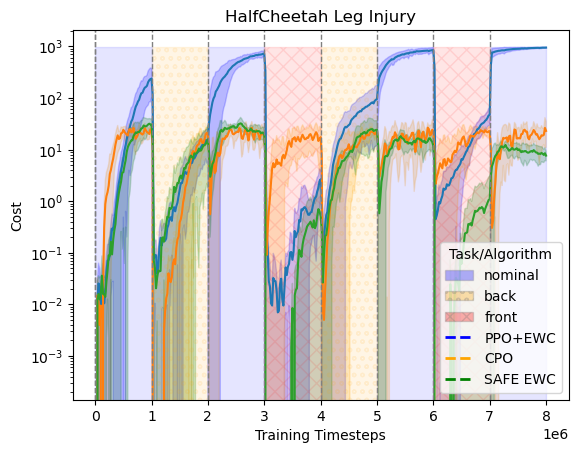} 
        \caption{Training cost curves.}
        \label{fig:cheetah_cost}
    \end{subfigure}
    
    \caption{Rewards and costs during training with task changes for the HalfCheetah environment. The tasks, shown by the background color, correspond to the tasks shown in Fig. \ref{fig:task_sequence}.}
    \label{fig:cheetah_train}
\end{figure}

First, we can inspect the rewards throughout the training process. Fig. \ref{fig:cheetah_reward} shows the rewards for each algorithm across task changes. When a task changed, there was a dramatic difference in reward, highlighting the unique challenges of each task. By viewing this figure task-by-task, we can qualitatively assess the catastrophic forgetting of each algorithm. In all cases, CPO, the agent trained with the non-continual RL algorithm, appeared to forget. Its reward was lower than the final reward of the last time it experienced the task. At first, the PPO+EWC and Safe EWC agents also show this behavior. However, as the task was revisited, the agents trained using EWC methods forgot less. In the back task, all agents forgot their policy, but the CPO agent had the largest reward drop since it converged quicker than the other algorithms. In the front task, the agents obtained using the EWC methods experienced minimal forgetting.

Next, we can inspect the costs throughout the training process, shown in Fig. \ref{fig:cheetah_cost}. Here, we can clearly see the advantage of CPO over PPO+EWC. The PPO+EWC agent completely ignored the velocity constraint and went as fast as possible to maximize reward. On the nominal task, the Safe EWC agent had higher costs than the CPO agent at first. However, it was significantly safer than the PPO+EWC agent. Additionally, by the time the nominal task was experienced for the fourth time, the Safe EWC agent violated safety less than the CPO agent. This indicates that Safe EWC may encourage the agent to remember safety constraints across task visits. In the back and front tasks, the Safe EWC and PPO+EWC agents had around the same or fewer safety violations than the CPO agent. This is likely due to the poorer task performance (see Fig. \ref{fig:cheetah_reward}), leading to fewer chances of violating safety.

\begin{figure}
    \centering
    \includegraphics[width=\columnwidth]{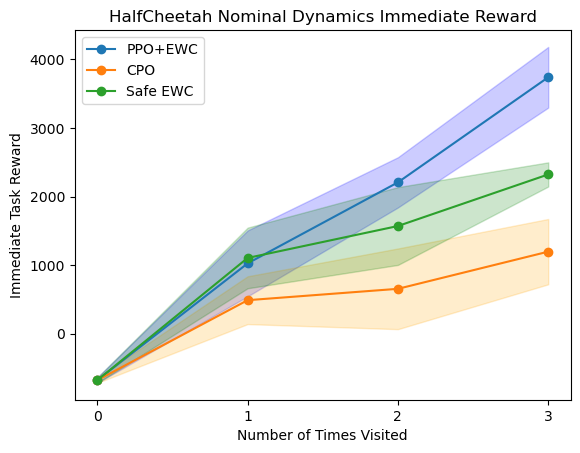}
    \caption{Immediate reward when experiencing nominal dynamics for the HalfCheetah. This measures how well the policy under nominal conditions is remembered.}
    \label{fig:immediate_reward}
\end{figure}

Further evidence for the strengths of each algorithm from a continual reinforcement learning perspective is shown in Fig. \ref{fig:immediate_reward}. This figure shows the immediate reward when experiencing nominal dynamics in the task sequence. The PPO+EWC agent had the highest positive slope, indicating it remembered more each time it revisited the task. The CPO agent was more stagnant, but slightly improved at the end, showing the inability of safe RL algorithms to effectively avoid catastrophic forgetting. The Safe EWC agent is in between, demonstrating the ability for safe continual RL algorithms to balance the tradeoff between continual learning and safety.

Beyond qualitative training curves, we can compute metrics to determine the strengths and weaknesses of each algorithm. We consider the following metrics.

\begin{itemize}
    \item \textbf{Total Cost}: the average total cost across each individual task. Calculated for each task as follows where $N$ is the number of times a task is visited and costs is a vector of the costs for each training timestep.

    \begin{equation}
        \frac{1}{N} \sum_{i=1}^{\text{len(costs)}} \text{costs}_i
    \end{equation}
    \item \textbf{Task Forget Percentage}: the percentage drop in performance from the previous time a task was experienced to the next time it is experienced. Calculated for each task as follows where final is the reward the last time the task was visited and immediate is the first episodic reward the next time the task is seen.

    \begin{equation}
        100 \times \frac{1}{N} \sum_{i=1}^{N} \frac{\text{final}-\text{immediate}}{|\text{final}|}
    \end{equation}
    
    \item \textbf{Final Task Reward}: the final reward across all visits of each task, measuring asymptotic performance.
\end{itemize}

\begin{table}[t]
\caption{HalfCheetah task sequence performance metrics. Mean $\pm$ standard deviation across 5 seeds.}
\centering
\label{tab:halfcheetah}
\begin{tabular}{@{}llll@{}}
\toprule
Agent & Nominal            & Back             & Front          \\ \midrule
\multicolumn{4}{c}{Total Cost ($\downarrow$)}                                           \\ \midrule
CPO       & $787.4 \pm 112.1$     & $\mathbf{386 \pm 183.5}$  & $629.7 \pm 298.7$ \\
PPO+EWC   & $25043.7 \pm 3202$ & $1074 \pm 792.3$ & $707.1 \pm 91.8$ \\
Safe EWC  & $\mathbf{680.1 \pm 34}$     & $395.6 \pm 228.6$  & $\mathbf{17.4 \pm 19.9}$  \\ \midrule
\multicolumn{4}{c}{Task Forget Percentage (\%) ($\downarrow$)}                                     \\ \midrule
CPO       & $46.6 \pm 19.9$      & $67.9 \pm 22.5$    & $108 \pm 60$  \\
PPO+EWC   & $26.1 \pm 7.9$      & $71.8 \pm 7.1$    & $\mathbf{30.8 \pm 17.7}$  \\
Safe EWC  & $\mathbf{19.6 \pm 15.6}$      & $\mathbf{62.3 \pm 34.4}$    & $36.2 \pm 8.6$  \\ \midrule
\multicolumn{4}{c}{Final Task Reward ($\uparrow$)}                                           \\ \midrule
CPO       & $2034.3 \pm 190.1$     & $1611.5 \pm 93.9$  & $1122.9 \pm 242.1$ \\
PPO+EWC   & $\mathbf{4690.1 \pm 429.6}$ & $\mathbf{1706.9 \pm 342.9}$ & $\mathbf{1385.5 \pm 99.4}$ \\
Safe EWC  & $2692.6 \pm 161.5$     & $1634.3 \pm 117.7$  & $1143.6 \pm 248.5$  \\ \bottomrule
\end{tabular}
\end{table}

The quantitative metrics for the HalfCheetah task sequence, averaged across the five seeds, are shown in Table \ref{tab:halfcheetah}. First, we can consider the average total cost. Unsurprisingly, the PPO+EWC agent had much higher costs than the CPO and Safe EWC agents. In the nominal and front cases, the Safe EWC agent had less cost than the CPO agent. On the front task, the Safe EWC agent had lower reward (Fig. \ref{fig:cheetah_reward}) than the CPO agent. This means it traveled slower, violating the velocity constraint less. The Safe EWC agent may violate safety less in the nominal case because it can remember safe actions. In terms of task forget percentage, the agents obtained using the EWC methods forgot much less than the CPO agent. In the nominal and back tasks, the Safe EWC agent forgot less than the PPO+EWC agent. This may be because the Safe EWC agent converged quicker to a more stable, slower-moving policy that was more consistent or because safety was consistent across tasks. The final task reward of the Safe EWC and CPO agents are very similar, approaching the limits of distance that can be traveled without violating safety. However, the Safe EWC agent achieves a higher reward in the nominal task that was visited 4 times.

\subsection{Case Study 2: Ant}

\begin{figure}[t]
    \centering
    \begin{subfigure}[b]{\columnwidth}
        \centering
        \includegraphics[width=\columnwidth]{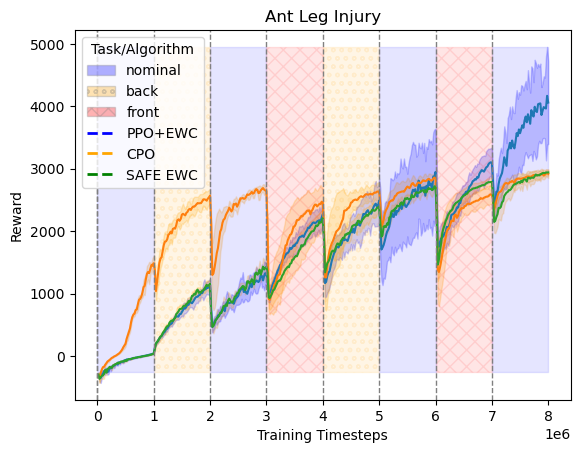} 
        \caption{Training reward curves.}
        \label{fig:ant_reward}
    \end{subfigure}

    \begin{subfigure}[b]{\columnwidth}
        \centering
        \includegraphics[width=\columnwidth]{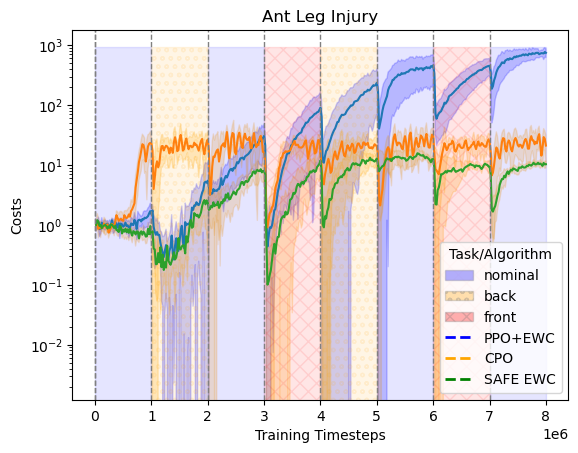} 
        \caption{Training cost curves.}
        \label{fig:ant_cost}
    \end{subfigure}
    
    \caption{Rewards and costs during training with task changes for the Ant environment. The tasks, shown by the background color, correspond to the tasks shown in Fig. \ref{fig:task_sequence}.}
    \label{fig:ant_train}
\end{figure}

We can perform a similar analysis for the Ant. Figure \ref{fig:ant_train} shows the training rewards and costs for the Ant task sequence. There are clear differences from the HalfCheetah scenario. For the first 4-5 million timesteps, the CPO agent was more sample-efficient and learned a reasonable policy much faster. The costs similarly reflected this. Neither the Safe EWC nor PPO+EWC agents could incur high costs because they did not move fast enough. However, the CPO agent forgot after each task switch. The PPO+EWC and Safe EWC agents exhibited positive backward transfer, improving on previous tasks by learning the next one. After the PPO+EWC and Safe EWC agents learned reasonable policies, the same cost relationship from the HalfCheetah case was held. The PPO+EWC agent ignored safety, and, in fact, the Safe EWC agent had significantly lower costs than the CPO agent. 

\begin{table}[t]
\caption{Ant task sequence performance metrics. Mean $\pm$ standard deviation across 5 seeds. The standard deviation was rounded to conserve space.}
\centering
\label{tab:ant}
\begin{tabular}{@{}llll@{}}
\toprule
Agent & Nominal            & Back             & Front          \\ \midrule
\multicolumn{4}{c}{Total Cost ($\downarrow$)}                                           \\ \midrule
CPO       & $873.6 \pm 78$     & $913.9 \pm 145$  & $872 \pm 34$ \\
PPO+EWC   & $11543.4 \pm 4563$ & $2551.1 \pm 2308$ & $6846.5 \pm 3216$ \\
Safe EWC  & $\mathbf{323 \pm 30}$     & $\mathbf{218.2 \pm 86}$  & $\mathbf{304.8 \pm 103}$  \\ \midrule
\multicolumn{4}{c}{Task Forget Percentage (\%) ($\downarrow$)}                                     \\ \midrule
CPO       & $15.1 \pm 12$      & $50.8 \pm 20$    & $44.4 \pm 18$  \\
PPO+EWC   & $-402.2 \pm 174$      & $-7 \pm 20$    & $\mathbf{21.2 \pm 13}$  \\
Safe EWC  & $\mathbf{-507.2 \pm 187}$      & $\mathbf{-16.7 \pm 7}$    & $31.9 \pm 13$  \\ \midrule
\multicolumn{4}{c}{Final Task Reward ($\uparrow$)}                                           \\ \midrule
CPO       & $2901.4 \pm 59$     & $\mathbf{2634.5 \pm 142}$  & $2590.2 \pm 152$ \\
PPO+EWC   & $\mathbf{3972.8 \pm 487}$ & $2407.2 \pm 320$ & $\mathbf{3103.1 \pm 228}$ \\
Safe EWC  & $2880.8 \pm 53$     & $2319.2 \pm 90$  & $2739.2 \pm 69$  \\ \bottomrule
\end{tabular}
\end{table}

These findings are reinforced by the quantitative metrics shown in Table \ref{tab:ant}. The Safe EWC agent had the lowest average total cost, as it maintained a lower cost than the CPO agent throughout training. As a consequence, the CPO agent obtained a higher final task reward. In reality, the algorithm used would depend on the risk tolerance of the operator. The PPO+EWC agent ignoring safety would likely be considered unacceptable. However, in this continual learning setting, the PPO+EWC and Safe EWC agents both showed significantly lower forgetting than CPO. For two tasks, they improved by learning on another task (hence the negative forgetting), i.e., positive backward transfer. In all cases, the CPO agent forgot. However, the CPO agent remembered the nominal Ant task much better than the nominal HalfCheetah task, improving from an average of 46.6\% forgetting to 15.1\% forgetting. This implies that the ability of an agent trained using a safe RL algorithm to avoid catastrophic forgetting can be task-dependent, warranting future studies.
\section{Discussion} \label{sec:discussion}

The results presented in Section \ref{sec:results} demonstrated the need for research in safe continual RL. We showed that an agent obtained using constrained policy optimization exhibited more catastrophic forgetting than one obtained using proximal policy optimization with elastic weight consolidation. However, the PPO+EWC agent ignored safety to maximize reward. We demonstrated that a simple modification to the reward function turns PPO+EWC into a continual RL algorithm which produces agents that compete with CPO in safety constraint satisfaction for the HalfCheetah and Ant velocity tasks under leg removal faults. However, this is a first attempt at studying the intersection of the fields of safe and continual RL. There are many sophisticated mechanisms in continual RL (experience replay buffers, knowledge bases, expansion-based approaches, etc.) that can be modified to remember safety. At the same time, the impact of non-stationarity on the safety of realistic systems needs to be analyzed more. It is not clear what types of tasks or task sequences require mechanisms from continual RL. 
\section{Conclusion} \label{sec:conclusion}

In this paper, we studied the intersection of safe and continual reinforcement learning. We evaluated the cost, reward, and percentage forgetting of agents trained using constrained policy optimization (CPO), a safe RL algorithm, proximal policy optimization with elastic weight consolidation (PPO+EWC), a continual RL algorithm, and a proposed modification to PPO+EWC called Safe EWC that shaped the reward to penalize costs. We evaluated these on the MuJoCo HalfCheetah and Ant environments with velocity constraints. Non-stationarity was modeled by removing the front or back limbs from the systems, emulating extreme damage to the system or a fault. We found that CPO agents maintained a low cost throughout learning but experienced catastrophic forgetting. We found that agents obtained using PPO+EWC experienced less catastrophic forgetting, remembering more each time a task was visited. Agents trained with Safe EWC, a proposed simple safe continual RL algorithm, maintained low total cost, low forgetting, and high task reward. However, the properties of the algorithms were influenced by the nature of the system and task. The PPO+EWC and Safe EWC agents were less sample efficient on the Ant, leading to positive backward transfer. The overall initial success of Safe EWC agents and unanswered questions about the influence of types of non-stationarity on safe RL algorithms call for future research in this field.

\bibliographystyle{ieeetr}
\bibliography{references}

\end{document}